\documentclass[letterpaper,10pt,conference]{ieeeconf}
\IEEEoverridecommandlockouts   
\usepackage[english]{babel}
\usepackage{amsmath}
\usepackage{lipsum} 
\usepackage[T1]{fontenc}
\usepackage{newtxtext}
\usepackage{tabularx}
\usepackage[cmintegrals]{newtxmath}
\usepackage{bm} 
\usepackage[linesnumbered,ruled,vlined]{algorithm2e}
\usepackage[utf8]{inputenc}
\usepackage{graphicx}
\usepackage{graphics}
\usepackage{subcaption}
\usepackage{siunitx}
\usepackage{multirow}
\usepackage{float}
\usepackage{caption}
\usepackage{gensymb}
\usepackage{hyperref}
\usepackage{epstopdf}
\usepackage{enumerate}
\usepackage{multicol,multirow}
\usepackage{booktabs}
\usepackage{cite}
\usepackage{balance}
\usepackage{tikz,xcolor}
\usepackage{academicons}
\DeclareCaptionLabelSeparator{periodspace}{.\quad}
\usepackage{svg}
\definecolor{orcidlogocol}{HTML}{A6CE39}
\graphicspath{{figures/}} 
\definecolor{lime}{HTML}{A6CE39}
\DeclareRobustCommand{\orcidicon}{%
	\begin{tikzpicture}
	\draw[lime, fill=lime] (0,0) 
	circle [radius=0.16] 
	node[white] {{\fontfamily{qag}\selectfont \tiny ID}};
	\draw[white, fill=white] (-0.0625,0.095) 
	circle [radius=0.007];
	\end{tikzpicture}
	\hspace{-2mm}
}

\foreach \x in {A, ..., Z}{
	\expandafter\xdef\csname orcid\x\endcsname{\noexpand\href{https://orcid.org/\csname orcidauthor\x\endcsname}{\noexpand\orcidicon}}
}

\definecolor{red}{rgb}{1.00,0.00,0.00}
\definecolor{blue}{rgb}{0.00,0.00,1.00}
\definecolor{green}{rgb}{0.30, 0.50,0.00}

\newcommand{\cblue}[1] {\textcolor{blue}{#1}}

\begin{document}
\title{Robust and Dexterous Dual-arm Tele-Cooperation using \\ Adaptable Impedance Control}

\author{Keyhan~Kouhkiloui~Babarahmati, Mohammadreza~Kasaei, Carlo~Tiseo, Michael~Mistry and Sethu~Vijayakumar 
\thanks{This work is supported by EU H2020 project: Enhancing Healthcare with Assistive Robotic Mobile Manipulation (HARMONY, 101017008) and the Alan Turing Institute, UK.}
\thanks{Keyhan~Kouhkiloui~Babarahmati, Mohammadreza~Kasaei, Michael~Mistry and Sethu~Vijayakumar are with the School of Informatics, University of Edinburgh, UK and Carlo~Tiseo is with the School of Engineering and Informatics, University of Sussex, UK. {Email: keyhan.kouhkiloui@ed.ac.uk}}}

\maketitle

\begin{abstract}
In recent years, the need for robots to transition from isolated industrial tasks to shared environments, including human-robot collaboration and teleoperation, has become increasingly evident. Building on the foundation of Fractal Impedance Control (FIC) introduced in our previous work, this paper presents a novel extension to dual-arm tele-cooperation, leveraging the non-linear stiffness and passivity of FIC to adapt to diverse cooperative scenarios. Unlike traditional impedance controllers, our approach ensures stability without relying on energy tanks, as demonstrated in our prior research. In this paper, we further extend the FIC framework to bimanual operations, allowing for stable and smooth switching between different dynamic tasks without gain tuning. We also introduce a telemanipulation architecture that offers higher transparency and dexterity, addressing the challenges of signal latency and low-bandwidth communication. Through extensive experiments, we validate the robustness of our method and the results confirm the advantages of the FIC approach over traditional impedance controllers, showcasing its potential for applications in planetary exploration and other scenarios requiring dexterous telemanipulation. This paper's contributions include the seamless integration of FIC into multi-arm systems, the ability to perform robust interactions in highly variable environments, and the provision of a comprehensive comparison with competing approaches, thereby significantly enhancing the robustness and adaptability of robotic systems.
\end{abstract}

\IEEEpeerreviewmaketitle

\section{Introduction} \label{sec: Introduction}
Robots are valuable resources for adaptable and interchangeable manufacturing. The advancements in technology brought about by these robots significantly contributed to the enhancement of our overall well-being and led to significant changes in our population. Currently, with the aging population, we are experiencing a decline in the number of people in the workforce and an escalating need for healthcare services catering to age-related ailments. Robots have the potential to address these demands by facilitating improved healthcare, minimizing work-related injuries, and decreasing risks for operators in hazardous environments \cite{Li2018,tommasino2017,tommasino2017task,tiseo2018bipedal}. The successful execution of these tasks requires the presence of control frameworks that can swiftly transition between various tasks within seconds, requiring minimal involvement from a user or operator. This is particularly important because the operator might lack the necessary technical expertise to reprogram the robot's controller in such situations.

\begin{figure}[t]
\centering
\begin{subfigure}[b]{0.99\columnwidth}
\includegraphics[width=0.99\columnwidth, trim=0.0cm 21cm 0.0cm 0.0cm,clip]{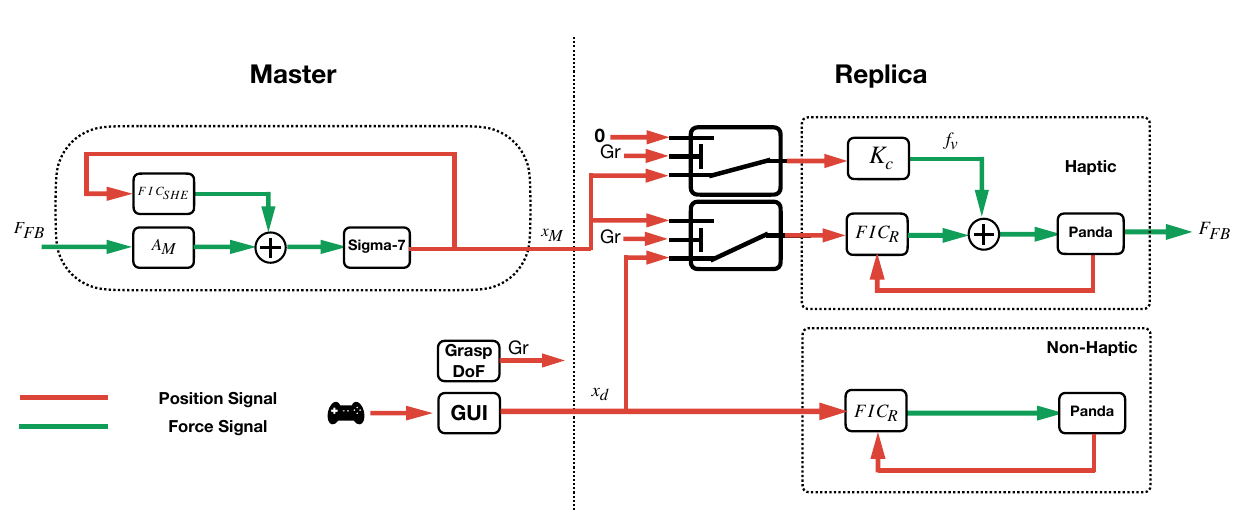}
\caption[]{}
\label{fig:proposedCtrlBlocks}
\end{subfigure}
\begin{subfigure}[b]{0.9\columnwidth}
\centering 
\includegraphics[width=0.9\columnwidth]{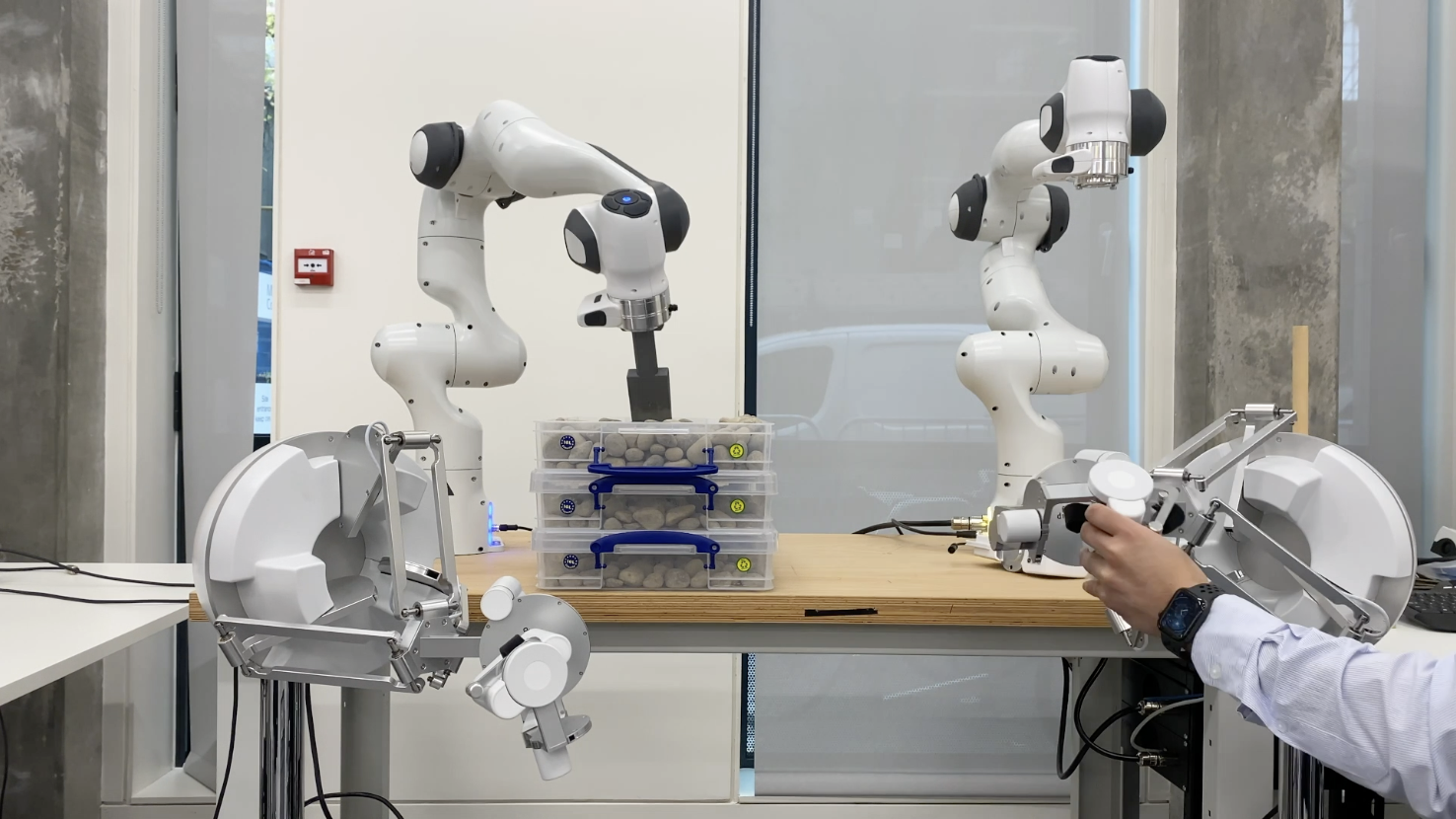}
\vspace{-1mm}
\caption[]{} 
\label{fig:dual_arm_teleCoop_setup}
\end{subfigure}
\caption{(a) \textbf{Haptic}: The master moves the replica (Panda, Franka Emika AG) by applying a virtual force ($f_\text{v}$). The interaction force/torque feedback at the end-effector ($F_{\text{FB}}$) is scaled in A$_\text{M}$ to generate the haptic feedback. The operator can also act on the replica end-effector's reference pose ($x_\text{d}$) by activating the master device with its grasp joint or via the GUI. \textbf{Non-Haptic}: The replica is controlled by issuing sequences of $x_\text{d}$ that act as via points for the trajectory of the replica. (b) Dual-arm Tele-cooperation Setup.}
\vspace{-3mm}
\end{figure}

Multiple solutions have been proposed to deploy robots in the applications mentioned above that we classify in four categories: high-level algorithms, adaptable controllers, smart sensing, and adaptable mechatronics \cite{braun2013robots,khoramshahi2020,nakanishi2011stiffness,Kronander2016,Tadele2014,Averta2020,hogan2018impedance}. However, it is essential for all these components to work together in order to make robots sufficiently flexible and robust for integration into our daily lives. The high-level algorithms category encompasses various optimization and machine learning methods that are employed to plan and adjust the actions of the robot. Adaptable controllers refer to architectures that enable a certain level of adaptation by integrating the action plan with sensor information. Smart sensing methods utilize multiple sensors to accurately estimate and predict the states of both the robot and the environment. Adaptable mechatronics involves technologies that can adapt to the environment, either through passive means such as soft materials and composites, or through active means such as Variable Stiffness Actuators and smart materials. While a comprehensive solution would require integrating technologies from all these fields, the analysis presented in this paper primarily focuses on adaptable controllers, with some consideration given to their integration capabilities with other technologies.

Interaction controllers are based on the Port-Hamiltonian frameworks, and they either rely on admittance or impedance control architectures \cite{Hogan1985,Hogan2014,hogan2018impedance}. Impedance controllers control the force based on the tracking accuracy. Admittance controller trade-off the robot tracking accuracy based on the interaction's desired force \cite{Hogan1985}. When it comes to these controllers, a general guideline suggests that the impedance controller is advantageous when seeking robust interaction in an unfamiliar environment, and it does not necessitate the use of a force sensor. Conversely, admittance controllers tend to perform better under controlled interaction conditions, such as when there is a clearly defined interaction point and a force/torque sensor is available, particularly in its commonly used implementation \cite{Erden2015,Li2018,Yang2011,babarahmati2020,Babarahmati2022}. One notable limitation of these admittance controllers is their lack of back-drivability when the interaction bypasses the force/torque sensor. As a result, they are most suitable for non-redundant mechanisms where the null-space cannot be utilized to accommodate undesired motion that does not directly involve the end-effector, such as in cases of single joint actuation.

Either admittance or impedance controllers have been integrated into optimised controlled architectures and/or coupled with learning algorithms to implement adaptable controllers \cite{Averta2020,Erden2015,Li2018,braun2013robots,ferraguti2015}. Although these methods successfully implemented proof of concept applications, they are susceptible to lack of accurate models, misrepresentation of the interaction conditions, numerical instability, sensor noise and information delay  \cite{Kronander2016,lee2010passive}. Furthermore, they are task-dependent, computationally expensive, and require extensive task-specific parameters tuning, limiting their flexibility \cite{Babarahmati2022,babarahmati2020}. In contrast, animals are capable of dexterous robust interaction in challenging environment despite their bio-signals are noisy and transferred with substantial delays to withstand adverse conditions. Furthermore, they are also capable of learning, adapt, and transfer motor skills, which have fuelled the interest of both the robotics and neuro-scientific communities for the last few decades \cite{tiseo2020,tommasino2017,tommasino2017task,tiseo2018bipedal}. 

In this research, a dual-arm tele-cooperation setup, as depicted in \autoref{fig:dual_arm_teleCoop_setup}, is utilized. This setup enables the accomplishment of diverse tasks with minimal intervention from the user/operator, thanks to the implementation of the studied control architecture proposed in this study can be seen in \autoref{fig:proposedCtrlBlocks}. The control architecture presented here enables seamless teleoperation, manipulation, and human-robot collaboration, effectively eliminating the requirement for time-consuming parameter tuning. Building upon a previously developed architecture for dexterous teleoperation \cite{babarahmati2020}, this proposed method serves as an extension that has demonstrated robustness in dealing with information delays and reduced communication bandwidth. More specifically, this extension includes the addition of interfaces for trajectory transfer, enhancements in adjusting the impedance characteristics of the controller, and scaling the architecture to facilitate bi-manual tasks. In summary, the proposed method offers the following contributions:
\begin{enumerate}[i)]

\item Unlike traditional impedance controllers that are often designed for specific tasks, the proposed controller allows for a seamless transition between teleoperation and manipulation. Furthermore, it enables the coordination of multiple arms using synchronous via-points, a feature not commonly found in conventional controllers.

\item An important feature of the proposed controller is its ability to superimpose and coordinate independent FICs. By leveraging the conservative energy of the FIC, the proposed controller ensures the stability of the superimposition of independent controllers. This is a significant departure from traditional impedance controllers, which often require intricate design and tuning to ensure stability in multi-controller scenarios.

\item The proposed controller, building on the foundational work of FIC, showcases robustness in highly variable environments. This adaptability, especially in scenarios involving human interaction, sets it apart from traditional controllers that might require re-tuning or reconfiguration when faced with environmental changes.
\end{enumerate}

In essence, the proposed method brings about stable task switching without requiring gain adjustments and demonstrates versatile performance in multi-arm configurations, as supported by experimental validations involving drilling tasks and human-robot interactions.

This manuscript is organsied as follows: \autoref{sec: FIC} provides some preliminaries on fractal impedance control, \autoref{sec: TeleCoop_FIC} introduces the architecture of proposed controller, \autoref{sec: experimental_design} describes the carried out experiments, \autoref{sec: results} presents the results that are later discussed in \autoref{sec: discussion} and, finally, \autoref{sec: conclusion} draws the conclusions.  

\section{Preliminaries on the FIC} \label{sec: FIC}
The Fractal Impedance Controller (FIC) is a passive asymptotically stable controller having smooth autonomous trajectories \cite{Babarahmati2022,babarahmati2020}. It relies upon an anisotropic impedance to drive the system, which generates a conservative field rendering the controller stable to communication delays and reduced control bandwidth \cite{Babarahmati2022,babarahmati2020}. The conservative energy of the controller is also capable of getting stacked in series and in parallel multiple controllers and perform online tuning without affecting stability \cite{Babarahmati2022,babarahmati2020,tiseo2020,tiseo2020Planner,tiseo2021,tiseo2020bio}.

The controller's implemented for this work relies on the non-linear stiffness based on the model presented in \cite{Babarahmati2022}. 
\begin{equation} \label{Impedance}
	K\left(\tilde{x}\right)=\mathrm{Diag}\left(\begin{array}{lr}
	K_\text{d}\left(\tilde{x}_i\right)=K_{\zeta i}+K_{\nu i}\left(\tilde{x}_i\right),& \text{Div}\\\\
	 \left( \cfrac{4}{\tilde{x}^2_{i|\text{max}}}\right) 	\displaystyle{\int_0^{\tilde{x}_{i|\text{max}}}} K_\text{d}\left(\tilde{x}_i\right)\tilde{x}_i~d\tilde{x}_i ,& \text{Conv}\\
  	\end{array}\right)
\end{equation}
where $\tilde{x}=x_\text{d}-x$ is the pose error, $x_{\text{d}}$ is the desired state, $K_{\zeta i}$ is a constant stiffness for the $i^\text{th}$ task-space degree of freedom, $\tilde{x}_{i|\text{max}}$, $K_{\nu i}\left(\tilde{x}_i\right)$ are variable stiffness profile and the maximum displacement reached during the divergence (Div) respectively. The $K_\nu\left(\tilde{x}_i\right)$ used in this work is:

\begin{equation} 
\label{eq:Kv}
    \begin{array}{l}
    K_{\nu i}= \begin{cases}
	\cfrac{F_i|\text{max}}{|\tilde{x}_i|} - K_{\zeta i}, \  	& \ \text{if} \ |\tilde{x}_i| > x_{i|\text{b}} \\\\
	\exp{\left(\beta_i\tilde{x}_i^2\right)}, & \  \text{otherwise}
	\end{cases}\\
	\end{array}
\end{equation}
where $F_{i|\text{max}}$ is the maximum force or torque with respect to physical limitation of robots, $x_{i|\text{b}}$ is the pose error where the force/torque saturate at $F_{i|\text{max}}$, and $\beta_i=\sqrt{\ln \left(\left(F_{i|\text{max}}/x_{i|\text{b}}\right)-K_{\zeta i}\right)}/\left(x_{i|\text{b}}^2\right)$. Thus, the controlled system has the following dynamic equation at the end-effector:
\begin{equation}
\Lambda \ddot{x} -D \dot{x} + K(\tilde{x}) \tilde{x}=0
\end{equation}
where $\Lambda$, $D$ are the Cartesian inertia, positive damping respectively, which is obtained by the following joint torque control signal:
\begin{equation}
    \label{controlcmd}
    \begin{array}{ll}
    \tau_\text{c}&= J^T\left(- D \dot{x} + K(\tilde{x}) \tilde{x} \right)+ F_\text{ND}+ \left(I - J^{T}(JJ^T)^{-1}J \right) {\tau}_{\text{null}}=\\\\
    &=J^Th_\text{e}+ \left(I - J^{\dagger}J\right) {\tau}_{\text{null}}
    \end{array}
\end{equation}

\noindent where $J$ is the Jacobian, $F_\text{ND}$ is the compensation of the non-linear dynamics, $M$ is the inertia matrix and $\tau_\text{null}$ is the null-space torque. 
\section{TeleCoop-FIC: A Novel Approach to Dual-Arm Tele-Cooperation} \label{sec: TeleCoop_FIC}
The TeleCoop-FIC (\textit{Tele-Cooperation using Fractal Impedance Control}) exploits the teleoperation architecture introduced in \cite{babarahmati2020}, based on the architecture in \autoref{fig:proposedCtrlBlocks}. This method has been validated for haptic teleoperation with delays up to \SI{1}{\second} and communication bandwidth as low as \SI{20}{\hertz} between Master(M) and Replica(R) \cite{babarahmati2020}. The experimental setup was composed by a Panda Arm (Franka Emika) as Replica(R) robot, and a Sigma.7 Haptic Interface (Force Dimension) as Master(M) device. The replica is also equipped with a force/torque sensor at the end-effector to measure the interaction force. The previous work in \cite{babarahmati2020} showed dexterous interaction when dealing with a variegate set of environmental conditions without re-tunning. This work here extends more results and builds on this architecture to control multiple arms, and use the controller for manipulation, teleoperation, and cooperation with a human operator in multiple tasks. By exploiting the non-linear impedance profile and the stability properties of the FIC, we show that there is no need of re-tuning the proposed controller gains. 
\subsection{Haptic Teleoperation}
The TeleCoop-FIC exploits the teleoperation architecture, validated for haptic teleoperation with delays up to 1 second and communication bandwidth as low as 20 Hz between Master (M) and Replica (R). By leveraging the conservative energy of the FIC, we have explored the possibility of superimposing and coordinating independent FICs via the synchronization of their desired states. This property ensures the stability of the superimposition of independent controllers, allowing for seamless control of multiple arms.
\subsubsection{Master's Controllers}
The master wrench command for the Sigma-7's controller is sum of the scaled end-effector force ($A_\text{M}$) and the additional impedance controller -- FIC -- for Safety and Haptic Enhancement ($\text{FIC}_{\text{SHE}}$):
\begin{equation} \label{eq:fractalImpCtrlMasterSide}
    F_\text{M} = \text{FIC}_\text{SHE} + \text{A}_\text{M}=K_\text{M}\left(\tilde{x}_\text{M}\right)\tilde{x}_\text{M} - D_\text{M}\dot{x}_\text{M} + K_\text{A}F_\text{FB}
\end{equation}
\noindent where $K_{\text{M}}(\tilde{x})$ is a state-dependant stiffness matrix, $D_{\text{M}}$ is the damping, $F_\text{FB}$ is the measured wrench at the replica's end-effector, and $K_A=1$ is the scaling factor.

\subsubsection{Replica's Controller}
The replica's controller is the combination of the virtual force from the master ($J_\text{R}^{T}K_\text{c}x_\text{M}$), and the FIC$_\text{R}$ when the grasp DoF is pressed.
\begin{equation} \label{eq:fractalImpCtrlSlaveSide}
\begin{array}{ll}
\tau_{\text{FICr}} &=  J_\text{R}^{T}\left(K_\text{R}\left(\tilde{x}_\text{Rm}\right)\tilde{x}_\text{Rm} - D_\text{R}\dot{x}_\text{R}\right) + F_\text{ND}+ \\ & +\left(I - J^{\dagger}_\text{R}J_\text{R}\right) {\tau}_\text{Rnull}
\end{array}
\end{equation}
where $\tilde{x}_\text{Rm} = x_M - x_R$. On the other hand when the grasp DoF is not selected the equation is as follows:
\begin{equation} \label{eq:fractalImpCtrlSlaveSide_b}
\begin{array}{ll}
\tau_{\text{FICr}}&= J_\text{R}^{T}K_\text{c}x_\text{M} + J_\text{R}^{T}\left(K_\text{R}\left(\tilde{x}_\text{Rd}\right)\tilde{x}_\text{Rd} - D_\text{R}\dot{x}_\text{R}\right)+ \\ & + F_\text{ND} +\left(I - J^{\dagger}_\text{R}J_\text{R}\right) {\tau}_\text{Rnull}
        \end{array}
\end{equation}
where $\tilde{x}_\text{Rd} = x_d - x_R$.

\subsection{Non-Haptic Teleoperation}
\subsubsection{Master's Controller}
The master wrench command in this case is through the GUI that gives the operator the capability of setting new set-points graphically and the robot gets to any new set-point using Polynomial Path algorithm (cubic) along the desired line connecting the previous- and new set-point.

\subsubsection{Replica's Controller}
The replica's controller in this case is only the fractal impedance control ($\text{FIC}_{\text{R}}$):
\begin{equation} \label{eq:fractalImpCtrlSlaveSide_NH}
\begin{array}{ll}
       \tau_{\text{FICr}}&= J_\text{R}^{T}\left(K_\text{R}\left(\tilde{x}_\text{R}\right)\tilde{x}_\text{R} - D_\text{R}\dot{x}_\text{R}\right)+\\\\ &+ F_\text{ND}+\left(I - J^{\dagger}_\text{R}J_\text{R}\right) {\tau}_\text{Rnull}
\end{array}
\end{equation}

\subsection{Multi-Arm Implementation}
The TeleCoop-FIC enables easy switching from teleoperation to manipulation, and multiple arms can be easily coordinated by issuing synchronous via-points. The passive nature of the controller allows for switching between control modes without the need for tuning or changing its architecture. This capability has been validated in scenarios involving drilling various materials, both with and without human interaction. Recently, we have explored the possibility of superimposing and coordinating independent FICs via the synchronisation of their desired states \cite{tiseo2020,tiseo2020Planner,tiseo2021}. This property is rendered possible by the conservative energy of the FIC, which guarantees the stability of the superimposition of independent controllers. In other words, each FIC regards the others as a part of the environment, and its global asymptotic behaviour guarantees the stability of interaction \cite{Babarahmati2022}. It is worth noting that this does not ensure the success of the coordinated task, but it only guarantees that the involved robots will not behave erratically. 

However, our earlier experiments validated the superimposition either for redundant manipulator \cite{tiseo2020,tiseo2021} or coordination of a swarm of drones \cite{tiseo2020Planner}. The proposed method extends the experimental validation of the FIC superimposition properties to multiple redundant manipulators, allowing two arms to complete two bi-manual tasks during unknown interaction. The validation of these properties substantially simplifies the algorithmic complexity of the robot interaction controllers deployed in teleoperation and manipulation. Currently, multi-arm manipulation requires the simultaneous observation of the state of both arms through the grasp matrix, which couples the state of the two systems and increasing the algorithmic complexity \cite{minelli2019energy,lin2018projected}. In contrast, the proposed architecture allows solving the same problem just by considering the simplified dynamics generated by the FIC at the end-effector.

We have conducted extensive experimental validation of the FIC superimposition properties to multiple redundant manipulators, allowing two arms to complete bi-manual tasks during unknown interaction. By exploiting the non-linear impedance profile and the stability properties of the FIC, we demonstrate that there is no need for re-tuning the proposed controller gains. This robustness is further validated through experiments involving drilling, moving objects with unknown dynamics, and interacting with humans.

\section{Experimental Design} \label{sec: experimental_design}
The preliminary validation of the proposed architecture is presented in \cite{babarahmati2020}. The results show that the method enables dexterous haptic teleoperation in multiple scenarios with the minimum tuning of the controller parameters in order to compare the functionality of the proposed method with traditional impedance control. Furthermore, it is extremely robust to the presence of delays and reduced communication bandwidth between the master and the replica. The controllers' gains used in this manuscript are the same as the ones used for the experiments in \cite{babarahmati2020}. 

\subsection{Design of Experiments} \label{sec: design_of_experiments}
The experiments presented in this work have been designed to evaluate the performance of the TeleCoop-FIC in teleoperation, manipulation, and human-robot cooperation task, which were beyond and complement the previous work in \cite{babarahmati2020}. Particularly, we aim to verify that it can switch between tasks without an extensive and tedious process for adjusting the controller tuning. To do so, we looked at tasks involving a dynamic interaction such as drilling multiple materials and retaining movement accuracy when carrying a variable mass. The set-up used for the experiments consists of two Panda Arms and a Sigma.7 Haptic Interface. It shall be noted that the teleoperation experiments presented in this manuscript are performed in line-of-sight teleoperation. It shall be noted that the user is always controlling both the linear and the angular components of the end-effector pose. However, the adjustable boundaries of the FIC force saturation (\autoref{eq:Kv}) can be used to funnel the interaction in predetermined constraints, which will be exploited in one of the experiments described below to guide a human operator during a slanted drilling task. 

The first experiment (Exp.~1) compares the proposed method against a traditional impedance controller (IC) in a drilling manipulation task on a clamped wood brick. The task involved drilling a hole using a Dremel multi-tool. We compare the controller's ability to penetrate the wood when a similar command is issued and the generated interaction force. The IC control law used for the comparison is:

\begin{equation} \label{eq:IC}
          \tau_\text{IC}= J_\text{R}^{T}\left( K_{\zeta|\text{R}}\tilde{x}_\text{R} - D_\text{R}\dot{x}_\text{R}\right)+ F_\text{ND}+\left(I - J^{\dagger}_\text{R}J_\text{R}\right) \tau_\text{Rnull}
\end{equation}
where $K_{\zeta|\text{R}}$ is a constant stiffness and $D_\text{R}$ is the damping. 

The second, third, and forth experiments focus on evaluating the modularity of the proposed framework by performing drilling on multiple objects of: (a) Small wood brick, (b) Flat-surface plastic object, (c) Carton box, (d) Curved-surface plastic object and (e) Large wood brick with the support of an autonomous arm that is securing the work-piece.  For these experiments, we evaluate the accuracy and precision of drilling in a predetermined location in wood (Exp.~2), 3D printed PLA (Exp.~3) and a deformable carton box (Exp.~4).

The fifth experiment involves two scenarios of human-robot collaboration with the robot arms moving autonomously using the manipulation controller in the proposed architecture \autoref{fig:proposedCtrlBlocks}. The first scenario (Exp.~5.1) consists of the two arms are handling the work-piece, and the human is performing the drilling. To successfully execute the task, the two arms have to pick-up the work-piece, carry it near to the operator, supporting the piece during the drilling, and delivering it back to the initial position once the operator has finished the task. During the second scenario (Exp.~5.2), one of the arms is clamping the work-piece against the ground, while the second arm is holding the Dremel, and it is in charge of controlling the tool inclination during drilling by exploiting its non-linear impedance to generate virtual boundaries that funnel the drill towards the desired inclination. Meanwhile, the human controls the drilling action by pushing on the robot's terminal link.

The final experiment (Exp.~6) is the bi-manual manipulation of the box that is picked up, presented to the operator, filled by the operator and carried back to its initial position by the robot. This is to evaluate the robustness of the proposed method to change of the environmental state (e.g., object inertia) by evaluating its tracking accuracy while carrying a variable mass object (i.e., the box). 

\section{Results} \label{sec: results}
\begin{figure}[!htb]
\vspace{1mm}
\centering
\includegraphics[width=.9\columnwidth, trim=1cm 9.75cm 1.75cm 10.25cm,clip]{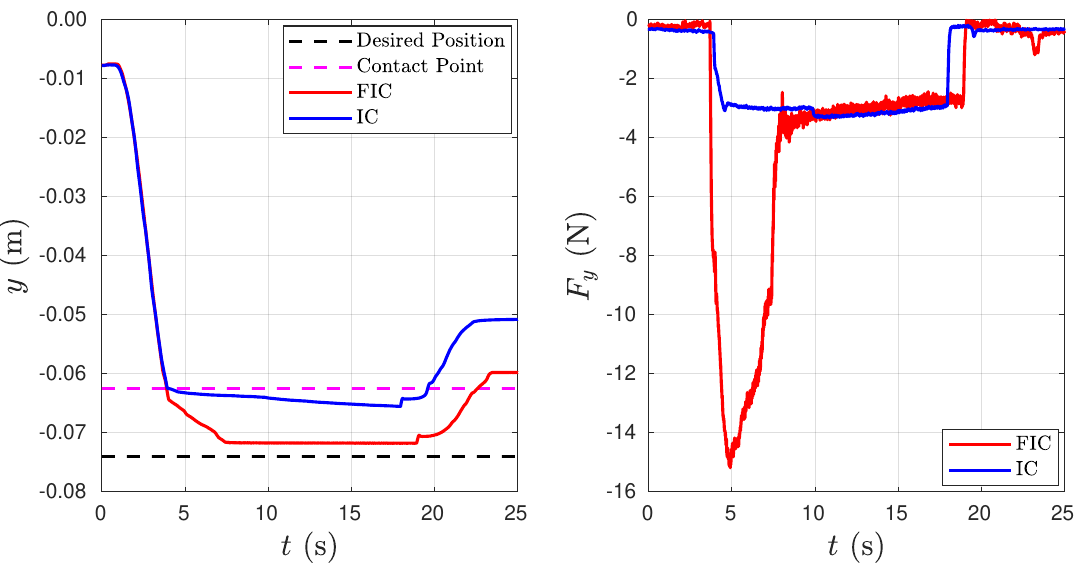}
\caption{The trajectory and the interaction force for the FIC and IC are compared when drilling a piece of wood.}
\label{fig:oneArm_drilling_IC_FIC}
\vspace{-7mm}
\end{figure}

The results from the first experiment show that the FIC controller is capable of generating an interaction force of \SI{15}{\newton} versus the \SI{3}{\newton} of the IC controller.  The higher force allows to get a penetration dept that is about \SI{1}{\centi\meter} deeper, reducing the final position error along the drilling direction ($\mathrm{y}$-axis) to about \SI{3}{\milli\meter} as shown in \autoref{fig:oneArm_drilling_IC_FIC}.  Such results are achieved without re-tuning the FIC controller or issuing a deeper desired position thanks to its nonlinear impedance that cannot be stabilised on a traditional IC, as reported in \cite{Babarahmati2022}. As a consequence, the FIC will have a stiffness similar to the IC for displacement within $x_\text{b}$ because it can exploit the nonlinear profile to generate high interaction force without requiring a large position error.  Thus, if a loss of contact occurs, the FIC will produce a smaller movement. To generate a similar behaviour with an IC controller would require variable impedance controllers that have more complex architecture and are only marginally stable.

\begin{figure}[!htb]
\centering
\includegraphics[width=.9\columnwidth, trim=16cm 9cm 10cm 9cm,clip]{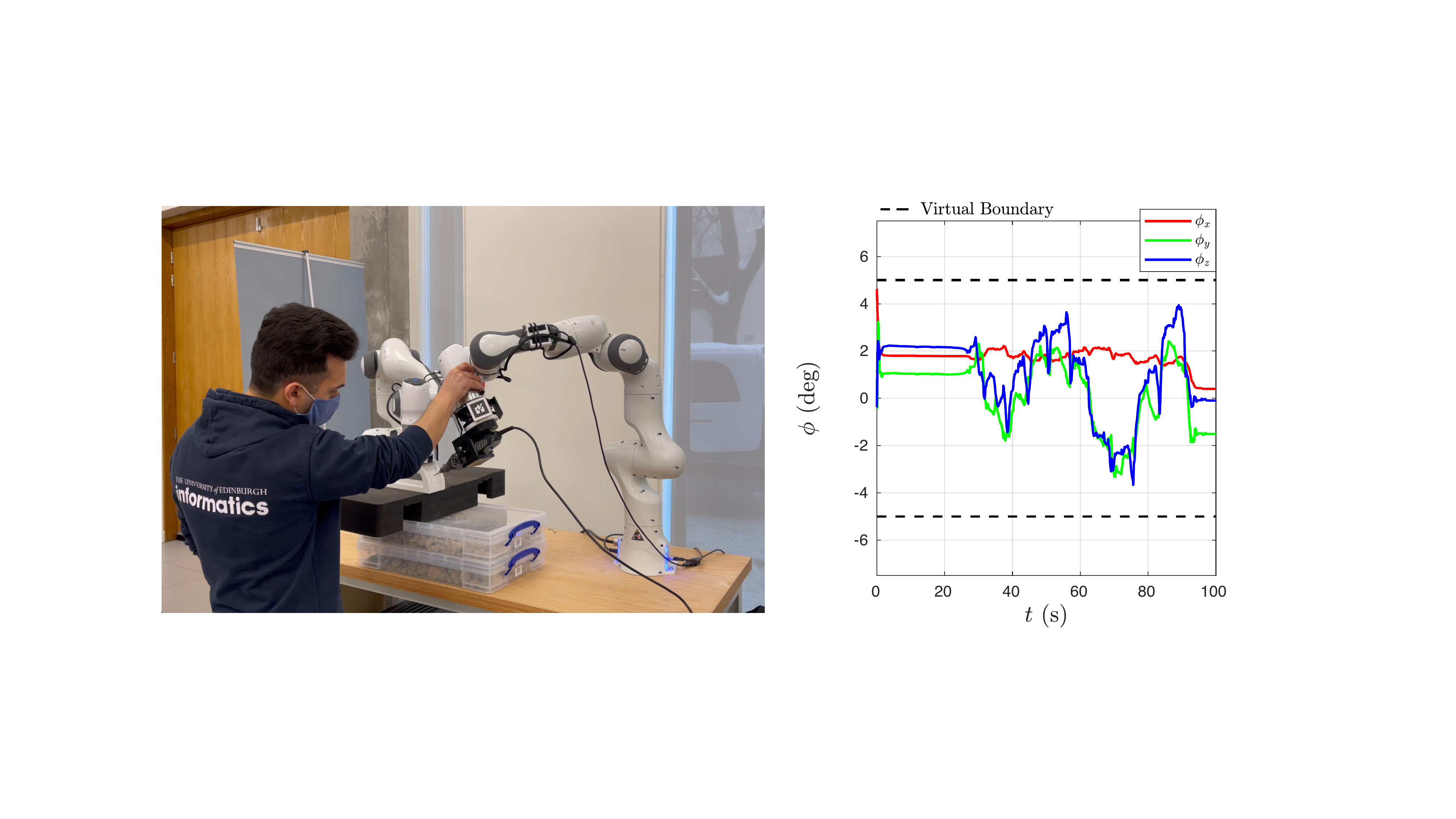}
\caption{ Experiment~5.1: Drilling on a curve surface with a constrained angular motion ($\pm 5\si{\degree}$).}
\label{fig:dual_arm_curved_drilling}
\vspace{-6mm}
\end{figure}

\begin{figure}[!htb]
\centering
\includegraphics[width=\columnwidth,trim=0.75cm 9.5cm 1.9cm 9.5cm, clip]{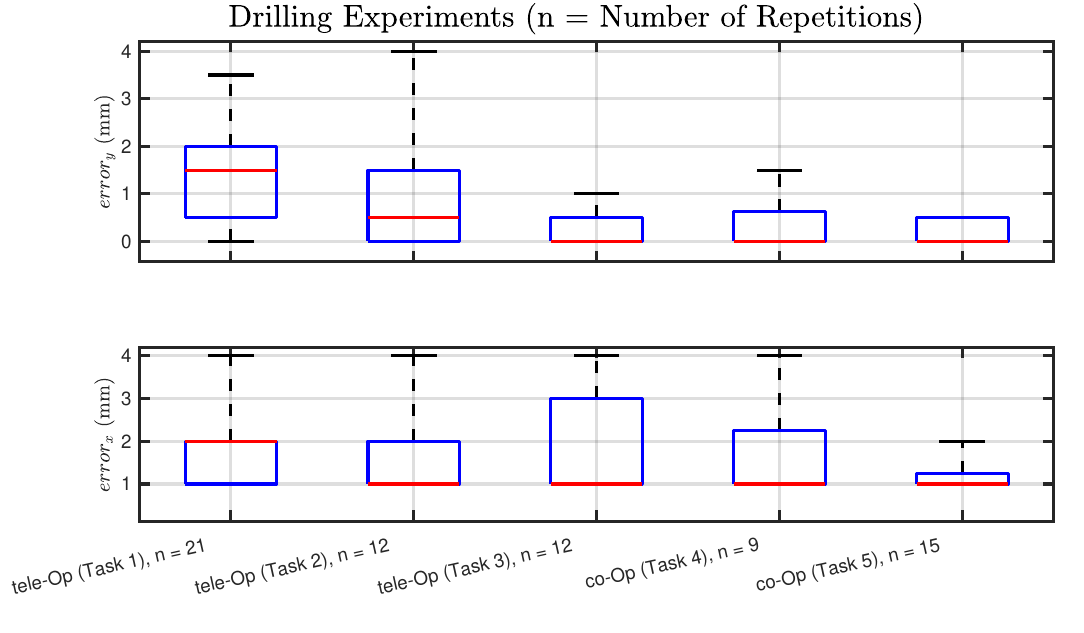}
\caption{Experiments 2 to 4: Teleoperation drilling of various objects. Experiment 5.1: Human-robot cooperation scenario 1 (Fig.~\ref{fig:dual_arm_curved_drilling}). Experiment 5.2: Human-robot cooperation scenario 2 (Fig.~\ref{fig:dual_arm_box_in_between_drilling}).}
\label{fig:pattern_drilling_table_boxplot}
\vspace{-4mm}
\end{figure}

The drilling position error for the teleoperation experiments are Exp 2, 3, and 4 in \autoref{fig:pattern_drilling_table_boxplot}. The mean errors indicate an accuracy between \SI{0.5}{\milli\meter} and \SI{1.5}{\milli\meter} that is mostly affected from the direction of the task on the $\mathrm{xy}$-plane rather than the material. The precision indicates that ranged from \SI{0.5}{\milli\meter} to \SI{1}{\milli\meter} depending on both direction and material. The average interaction forces measured during the three tasks depend from the material rigidity, and they are \SI{10}{\newton}, \SI{8}{\newton} and \SI{4}{\newton}, respectively. 

\begin{figure}[!htb]
\centering
\hspace{0mm}
\includegraphics[width=0.99\columnwidth, trim=25cm 5cm 1cm 7cm,clip]{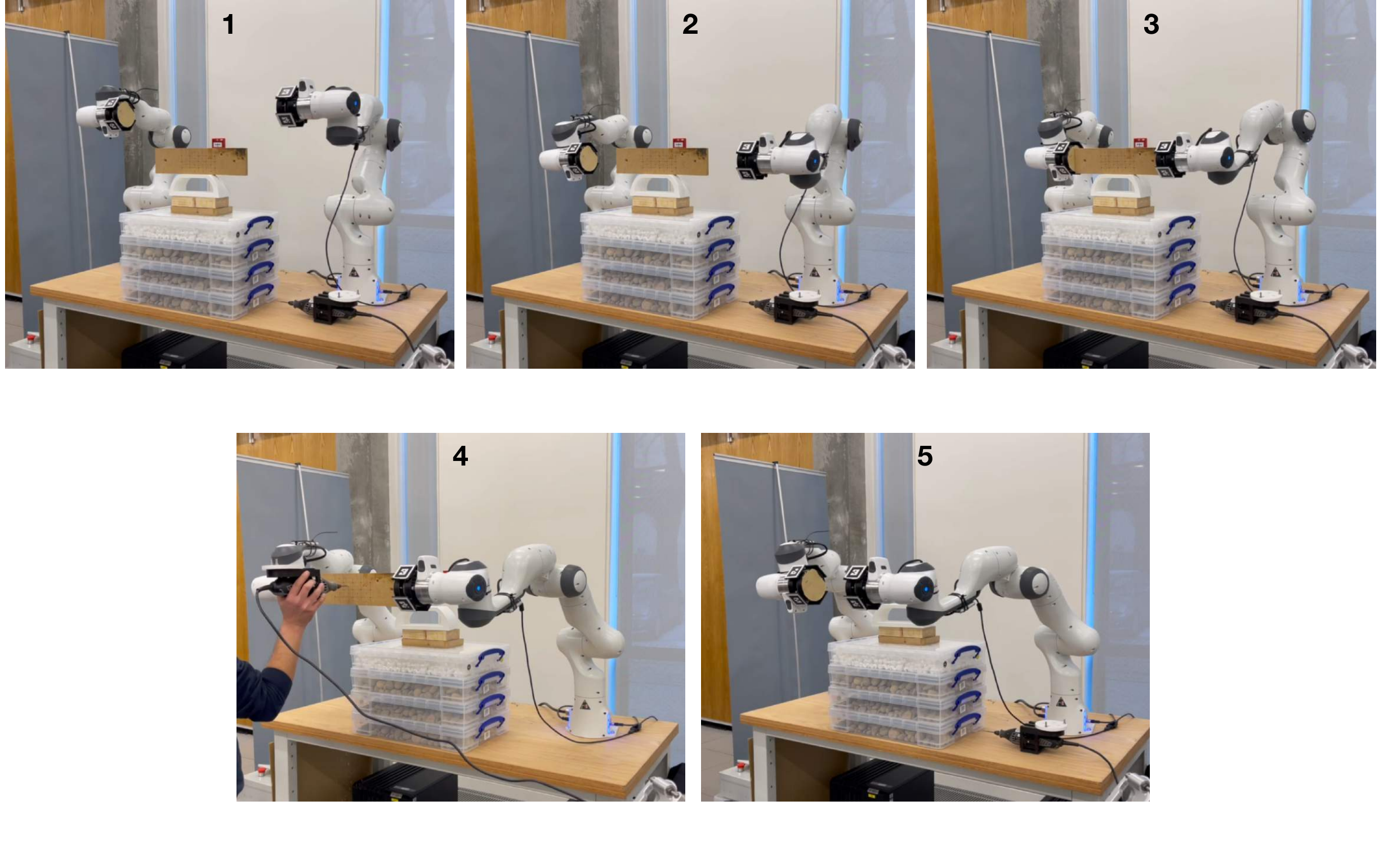} 
\vspace{-8mm}
\caption{Experiment~5.2: The robots start from via point 1 and go back to it. The object is picked up in 3. The drilling occurs in 4 and once the user has finished, the work-piece is removed in 5.}
\label{fig:dual_arm_box_in_between_drilling}
\vspace{-3mm}
\end{figure}

\begin{table*}[t]
\centering
\caption{Comparison of TeleCoop-FIC and Traditional Variable Impedance Controllers}
\resizebox{0.99\textwidth}{!}{
\begin{tabularx}{\textwidth}{lXX}
\toprule
\textbf{Feature/Property} & \textbf{TeleCoop-FIC} & \textbf{Traditional Variable Impedance Controllers} \\
\midrule
Task Transition & Seamless switching between teleoperation and manipulation. & Often requires reconfiguration for task switching. \\
\addlinespace
Multi-Arm Coordination & Coordination of multiple arms using synchronous via-points. & Limited multi-arm coordination capabilities. \\
\addlinespace
Stability & Uses conservative energy of FIC to ensure stability during superimposition of independent controllers. & Stability often requires intricate design and tuning, and mostly by running experiments to find the stability regions. \\
\addlinespace
Adaptability & Robust and compliant in highly variable environments without the requirement of controller gain re-tuning. & Might require controller gain re-tuning in changing environments and various scenarios. \\
\addlinespace
Energy Utilization & Exploits conservative energy properties of FIC, ensuring stability without additional energy tanks. & Might require additional energy tanks or mechanisms for ensuring stability by degrading the robot performance. \\
\addlinespace
Controller Modularity & Passive nature allows for easy switching between control modes without architecture changes. & Might require extensive controller reconfiguration for different modes. \\
\bottomrule
\end{tabularx}
}
\label{tab:TeleCoop_FIC_VIC_Comparison}
\vspace{-5mm}
\end{table*}

\begin{figure}[!htb]
\centering
\includegraphics[width=0.99\columnwidth, trim=25cm 5cm 1cm 7cm,clip]{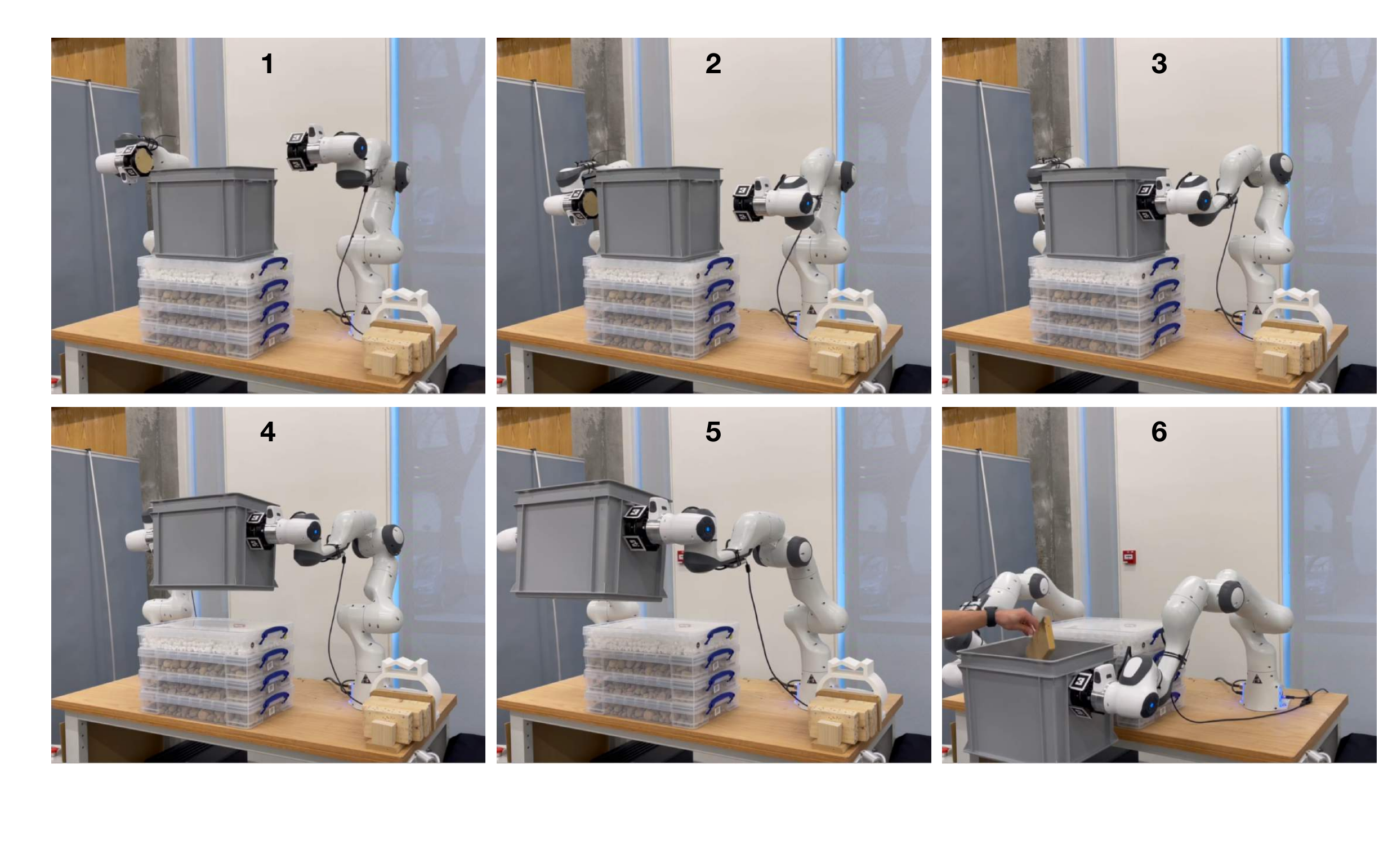} 
\vspace{-10mm}
\caption{Experiment~6: Snapshots of the robots reaching a box in 3 and lifting it up. They move it to 6 where it is loaded with \SI{2}{\kilo\gram} of materials. The robots finally bring it back to 3 before going back to 1.}
\label{fig:dual_arm_box_in_between_loading}
\vspace{-3mm}
\end{figure}

The error recorded from two cooperation experiments for drilling are the experiments in \autoref{fig:pattern_drilling_table_boxplot}. They indicate that we can expect sub-millimetre accuracy when the user is holding the drill. The experiment 5.1 proves that the two arm movement can be synchronised issuing coordinated via-points without relying on trajectory planning. The results described in \autoref{fig:dual_arm_box_in_between_drilling} indicate that it is possible to pick up the work-piece, carry it to the user, providing sufficient support to execute an accurate drilling, and retain stability and safety of interaction even in the event of a sudden loss of contact. The experiment 5.2 confirms that the robot can constrain the dremel inclination to the predetermined range of $\pm5\si{\degree}$ in each direction when drilling on a curve surface, as shown in \autoref{fig:dual_arm_curved_drilling}. The bi-manual manipulation task trajectories (Exp.~6) are shown in \autoref{fig:dual_arm_box_in_between_loading}. The arms start in the first via point and move to third to pick-up the box. The empty box is brought in to the 6$^\text{th}$ via point where the human loads the box with about \SI{2}{\kilo\gram} of material. Regardless of the added mass the robots retain a sufficient position accuracy that enables them to carry the box back to the 3$^\text{rd}$ via point, which is bounded by the FIC force saturation to a value that is $\leq \tilde{x}_\text{b}$. It is worth noting that the FIC is generating the maximum effort (i.e., forces and torques) selected for a task, and any position error greater than $\tilde{x}_\text{b}$ implies that the current load exceed the controller/robot capabilities. Thus, it will either require a change in the selected maximum effort in the controller, or a change of robot when the load exceed its mechanical capabilities. Experimental video is accessible through  \href{https://youtu.be/tHZ806Tjdb0}{{\cblue{https://youtu.be/tHZ806Tjdb0}}}.



\section{Discussion} \label{sec: discussion}

The proposed controller has been validated in various real-world scenarios, including drilling and human interaction. Its ability to handle these diverse tasks without extensive controller reconfiguration underscores its robustness compared to traditional impedance controllers. exploits the conservative energy properties of the FIC, ensuring stability even in challenging conditions. Traditional impedance controllers might require additional energy tanks or mechanisms to ensure stability, especially in scenarios with communication delays or reduced control bandwidth. It has also been validated in various real-world scenarios, including drilling and human interaction. Its ability to handle these diverse tasks without extensive controller reconfiguration underscores its robustness compared to traditional impedance controllers. \\
\indent The experiments show that the TeleCoop-FIC can easily switch from teleoperation to manipulation, and multiple arms can be easily coordinated by issuing synchronous via-points thanks to the FIC modularity. The modality of the architecture also allows to pass from manipulation to teleoperation simply by selecting the communication lines between the controller's modules in \autoref{fig:proposedCtrlBlocks}, enabled by the passive nature of the controller, as in-dept analysed in \cite{Babarahmati2022,tiseo2020,tiseo2020Planner}. It is capable of getting switched between control modes without the need of tuning or changing its architecture, but it can be done simply by selecting the information source using conditional statements. Additionally, the controller passivity also guarantees that the proposed method is robust to communication delays and reduced bandwidth \cite{babarahmati2020}. \\
\indent The tasks show that the method can exploit its nonlinear impedance to generate higher interaction forces and accurate tracking without need of accurate modelling of the interaction dynamics. This characteristics renders this method extremely robust because it does not requires the knowledge of the environmental dynamics to stabilise its behaviour. Therefore, it presents all of the benefits that constant impedance controllers have on variable impedance and admittance controllers, but it does not suffer their low accuracy \cite{Babarahmati2022,Averta2020,Li2018,minelli2019}. Such accomplishment is made possible by the tunable non-linear impedance profile that enables to select the required task precision, the maximum effort and/or constant stiffness that the controller produce. Moreover, the selection of these characteristics also upper-bounds the system maximum power and maximum kinetic energy as described in \cite{tiseo2020bio,tiseo2020Planner}. The nonlinear impedance allows to define low impedance funnels that can be used to guide the task performed by the user, as it has been shown in Exp.~5.2 when the user was drilling a curved work-piece. This capability might find application in training on new operators, guiding them on specific action patters, or respecting constraints that are otherwise difficult to respect (e.g. drilling with a constant angle on a curve surface). \\
\indent Furthermore, it is worth noting that the interaction with the robot during the task felt very natural and all the tasks required were successful at the first attempt, and the motor synchronisation was obtained only via the mechanical interaction between the robot and the users. Finally, to better understand the unique contributions and advantages of the TeleCoop-FIC, we present a direct comparison with traditional variable impedance controllers that can be seen in \autoref{tab:TeleCoop_FIC_VIC_Comparison}. TeleCoop-FIC can be considered for applications in manufacturing, space robotics, medical robotics and rehabilitation.
\section{Conclusion} \label{sec: conclusion}
\indent In Conclusion, the main observed limitation is that the task accuracy and precision should be in the order of millimetre based on our set-up, but it will vary based on the robots used. Especially considering that the tuning of the controller is straight forward and can be performed by non-technical operators, simply selecting the desired task precision ($\tilde{x}_\text{b}$), maximum effort ($F_\text{max}$) with respect to the physical limitation of the robot, and constant stiffness ($K_\zeta$). However, an identification of the maximum parameters is required to limit the parameter range to be within the mechanical capabilities of the robot. An identification procedure for this type of controller has been already developed, tested on multiple robots and is detailed in \cite{Babarahmati2022}. \\
\indent Future research should prioritize refining the controller's tuning identification procedure, making it adaptable across a diverse range of robot types. Emphasis should also be placed on enhancing the system's accuracy and precision, addressing the current millimeter-order precision limitations. Additionally, the system's vast potential in sectors like manufacturing, space robotics, medical robotics, rehabilitation, and training beckons dedicated exploration. 



\bibliography{main}
\bibliographystyle{IEEEtran}

\end{document}